\RequirePackage{etoolbox}
\csdef{input@path}%
{%
 {sty/},
 {img/},
}
\documentclass{elsevierbook}

\usepackage[numbers,sort&compress]{natbib}
\usepackage{graphicx}
\usepackage{siunitx}
\usepackage{amsmath}
\usepackage{amssymb}

\begin{document}

  
\Mainmatter
  \begin{frontmatter}

\chapter{Semantic Scene Segmentation for Robotics}\label{chap12}
\subchapter{Juana Valeria Hurtado and Abhinav Valada\footnote{Department of Computer Science, University of Freiburg, Germany, Emails: hurtadoj@cs.uni-freiburg.de, valada@cs.uni-freiburg.de }}


\begin{abstract}
Comprehensive scene understanding is a critical enabler of robot autonomy. Semantic segmentation is one of the key scene understanding tasks which is pivotal for several robotics applications including autonomous driving, domestic service robotics, last mile delivery, amongst many others. Semantic segmentation is a dense prediction task that aims to provide a scene representation in which each pixel of an image is assigned a semantic class label. Therefore, semantic segmentation considers the full scene context, incorporating the object category, location, and shape of all the scene elements, including the background. Numerous algorithms have been proposed for semantic segmentation over the years. However, the recent advances in deep learning combined with the boost in the computational capacity and the availability of large-scale labeled datasets have led to significant advances in semantic segmentation. In this chapter, we introduce the task of semantic segmentation and present the deep learning techniques that have been proposed to address this task over the years. We first define the task of semantic segmentation and contrast it with other closely related scene understanding problems. We detail different algorithms and architectures for semantic segmentation and the commonly employed loss functions. Furthermore, we present an overview of datasets, benchmarks, and metrics that are used in semantic segmentation. We conclude the chapter with a discussion of challenges and opportunities for further research in this area.
\end{abstract}

\begin{keywords}
\kwd{Semantic Segmentation}
\kwd{Scene Understanding}
\kwd{Visual Learning}
\kwd{Deep Learning}
\end{keywords}

\end{frontmatter}

\section{Introduction}
\label{sec1}
In order for robots to interact with the world, they should first have the ability to comprehensively understand the scene around them. The efficiency with which a robot performs a task, navigates, or interacts, strongly depends on how accurately it can comprehend its surroundings. Furthermore, the ability to understand context is crucial for safe operation in diverse environments~\cite{premebida2018intelligent}. However, accurate interpretation of the environment is extremely challenging, especially in real-world urban scenarios that are complex and dynamic. In these environments, robots are expected to perform their tasks precisely while they encounter diverse agents and objects. These environments themselves undergo appearance changes due to illumination variations and weather conditions, further increasing the difficulty of the task~\cite{valada2019self}.

\begin{figure}
\centering \centerline{
\includegraphics[width=0.99\linewidth]{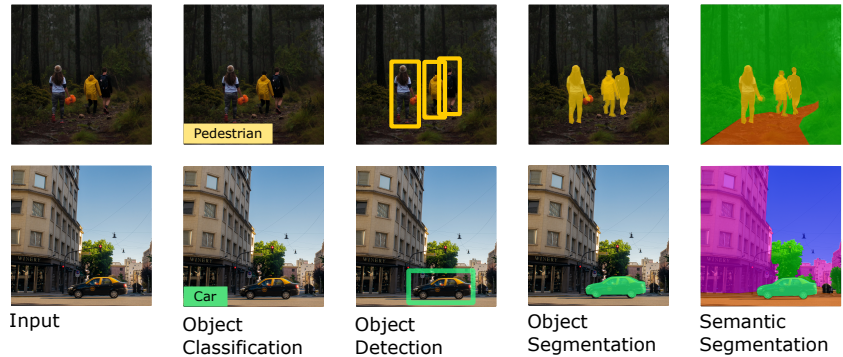}}
\caption{\it Each row shows an example with an input image and the corresponding output of different scene understanding tasks. Object classification identifies "what" objects compose the image, object detection predicts "where" the objects are located in the image, object segmentation outputs a mask that indicates the shape of the object. Semantic Segmentation further details the input image by predicting the label of all the pixels, including the background.}

\label{fig:perc_tasks}
\end{figure}

In robotics, scene understanding includes identifying, localizing, and describing the elements that compose the environment, their attributes, and dynamics. Research on novel techniques for automatic scene understanding has been exponentially increasing over the past decade due to the potential impact on numerous applications. The advances in deep learning as well as the availability of open-source datasets, and the increasing capacity of computation resources have facilitated the rapid improvement of scene understanding techniques~\cite{premebida2018intelligent}. These advances are most evident in image classification, where the goal is to determine what an image contains. The output of this classification task can be considered as a high-level representation of the scene which enables the identification of the various objects that are present by assigning a class label to them. A different mid-level capability, known as object detection, further details the scene by simultaneously classifying and localizing the objects in the image with bounding boxes. Although this task represents the scene in greater detail, it is still unable to provide essential information or object attributes such as the shape of objects. A closely related task, known as object segmentation, aims to fill this gap by additionally providing the shape of the object inside the bounding box in terms of their segmented boundaries. We present an overview of these perception tasks in Figure \ref{fig:perc_tasks}.

\begin{figure}
\centering \centerline{
\includegraphics[width=0.49\linewidth]{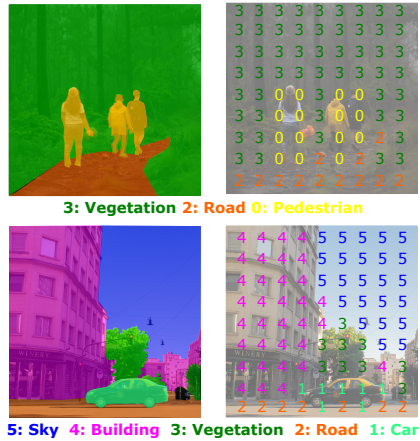}}
\caption{\it Illustration of the semantic segmentation output in which a semantic class label is assigned to each pixel in the image. The network predicts label indices for each pixel which is depicted with different colors for visualization purposes. The predictions overlaid on the input image is shown on the left and the label indices overlaid on the input image is shown on the right right.}
\label{fig:semsegcar}
\end{figure}

Semantic segmentation on the other hand presents a more integrated scene representation that includes classification of objects at the pixel-level, and their locations in the image. Semantic segmentation is considered as a low-level task that works at the pixel resolution and unifies object detection, shape recognition, and classification. By extending the previous tasks, semantic segmentation assigns a class label to each pixel in the image. Therefore, as we show in Figure~\ref{fig:semsegcar}, semantic segmentation models output a full-resolution semantic prediction that contains scene contexts such as the object category, location, and shape of all the scene elements including the background. Given that this prediction is performed for each pixel of the image, it is known as a dense prediction task.

Detailed understanding of the scene by assigning a class label to each pixel facilitates various downstream robotic tasks such as mapping, navigation, and interaction. This is an important step towards building complex robotic systems for applications such as autonomous driving~\cite{kalweit2020interpretable, radwan2020ijrr}, robot-assisted surgery~\cite{tewari2002technique, qin2020davincinet}, indoor service robots~\cite{boniardi2016autonomous, hurtado2021learning}, search and rescue robots~\cite{mittal2019vision}, and mobile manipulation~\cite{honerkamp2021learning}. Therefore, incorporating semantic information has strongly influenced several robotics areas such as Simultaneous Localization And Mapping (SLAM) and perception through object recognition and segmentation, highlighting the importance of semantics knowledge for robotics. Semantic information can be of special importance to tackle challenging problems such as perception under challenging environmental conditions. Robots operating in the real world encounter adverse weather, changing lightning condition from day through night. Accurately predicting the semantics in these conditions is of great importance for successful operation.

\begin{figure}
\centering \centerline{
\includegraphics[width=0.75\linewidth]{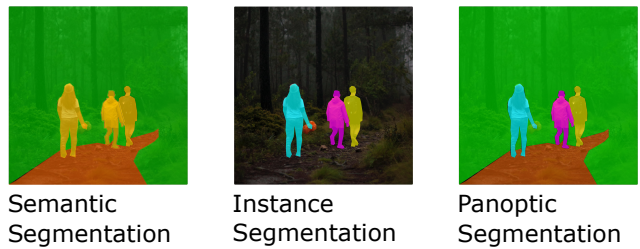}}
\caption{\it Comparison of semantic segmentation, instance segmentation, and panoptic segmentation tasks. Semantic segmentation assigns a class label to each pixel in the image and instance segmentation assigns an instance ID to pixels belonging to individual objects as well as semantic class label to each pixel in the image. The panoptic segmentation task unifies semantic and instance segmentation.}
\label{fig:multitask}
\end{figure}

In Semantic segmentation, we are not interested in identifying single instances, i.e., individual detections of objects in the scene. Therefore, the segmentation output does not distinguish two objects of the same semantic class as separate entities. Another closely related perception task known as instance segmentation, allows for distinguishing instances of objects of the same class in addition to pixel-level segmentation, and Multi-Object Tracking and Segmentation (MOTS)~\cite{voigtlaender2019mots} tackles the problem of obtaining temporally coherent instance segmentation. Furthermore, panoptic segmentation~\cite{kirillov2019panoptic} is a recently introduced task that unifies semantic and instance segmentation. We present a graphical comparison of these perception tasks in Figure~\ref{fig:multitask}. An additional task named Multi-Object Panoptic Tracking (MOPT)~\cite{hurtado2020mopt} further combines panoptic segmentation and MOTS. This elucidates the importance of developing models for a more holistic scene representation.

In this chapter, we present techniques for semantic scene segmentation, primarily using deep learning. We first discuss different algorithms and architectures for semantic segmentation, followed by the different loss functions that are typically used. We then discuss how different types of data and modalities can be used, and how video-classification models can be extended to yield temporal coherent semantic segmentation. Subsequently, we present an overview of datasets, benchmarks, and metrics that are used for semantic segmentation. Finally, we discuss different challenges and opportunities for developing advanced semantic segmentation models.

\section{Algorithms and Architectures for Semantic\\Segmentation}

The automatic understanding of the scene semantics has been a major area of research for decades. However, unprecedented advancement in semantic segmentation methods has only been achieved recently. Deep learning has played a significant role in enabling this capability, especially after the introduction of Fully Convolutional Networks~(FCNs)~\cite{long2015fully} which were proposed as a solution for semantic segmentation. In this section, we first provide a brief overview of semantic segmentation approaches used prior to Convolutional Neural Networks~(CNNs) and then present important deep learning approaches, their focus on improvements and limitations. We further present challenges and proposed solutions.

\subsection{Traditional Methods}

Typically, the traditional algorithms for image segmentation use clustering methodologies, contours, and edges information~\cite{weinland2011survey, sonka2014image}. Particularly for semantic segmentation, diverse approaches initially followed the idea of obtaining a pixel-level inference by considering the relationship between spatially close pixels. To do so, various features of appearance, motion, and structure ranging in complexity were considered including pixel color, surface orientation, height above camera~\cite{brostow2008segmentation}, histogram of oriented gradients (HOG)~\cite{dalal2005histograms}, Speeded Up Robust Features (SURF)~\cite{bay2008speeded}, amongst others. Approaches for image semantic segmentation range from simple thresholding methods in gray images to more complex edge-based approaches~\cite{lindeberg1997segmentation, barghout2014visual, osher2003geometric} or graphical models such as Conditional Random Fields (CRF) and Markov Random Fields (MRF)~\cite{ladicky2009associative, montillo2011entangled, yao2012describing}. Another group of approaches employ multiple individually pre-trained object detectors with the aim of extracting the semantic information from the image~\cite{ladicky2010and}. In general, as traditional approaches typically rely on a priori knowledge, such as the dependency between neighboring pixels, these methods require the definition of semantic and spatial properties with respect to the application to define segmentation concept. Moreover, these methods are limited to being able to segment a specific number of object classes which are defined by hand-selected parameters of the methods.

\subsection{Deep Learning Methods}

The introduction of deep learning methods and deep features presented important advances in computer vision tasks such as image classification, and led to the interest in using deep features for semantic segmentation. 
The initially proposed deep learning approaches for semantic segmentation used classification networks such as VGG~\cite{simonyan2014very} and AlexNet~\cite{krizhevsky2017imagenet}, and adapted them for the task of semantic segmentation by fine-tuning the fully connected layers~\cite{ning2005toward, ganin2014n, ciresan2012deep}. As a result, these approaches were plagued by overfitting and required significant training time. Additionally, the semantic segmentation performance was affected by the insufficient discriminative deep features that were learned by them.

Most subsequently proposed methods suffered from low performance, and consequently several refinement strategies were incorporated such as Conditional Random Fields (CRFs)~\cite{lafferty2001conditional}, Markov Random Fields (MRFs)~\cite{ning2005toward}, nearest neighbours~\cite{ganin2014n}, calibration layers \cite{ciresan2012deep}, and super-pixels~\cite{farabet2012learning, hariharan2014simultaneous}. Refinement strategies are still used as post-processing methods to enhance the pixel classification around regions where class intersections occur~\cite{ulku2019survey}.

\begin{figure}
\centering \centerline{
\includegraphics[width=0.6\linewidth]{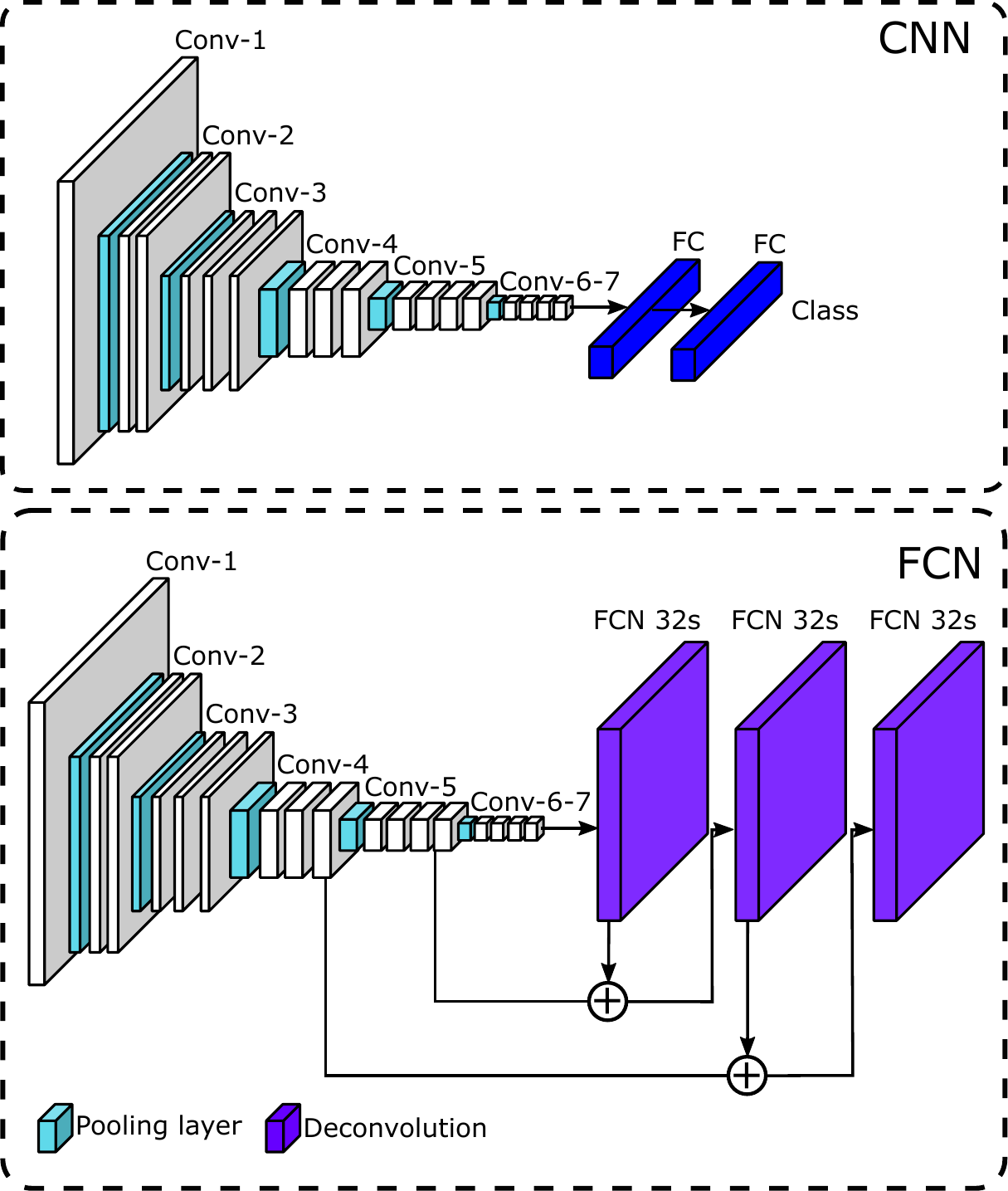}}
\caption{\it An example topology of a Convolutional Neural Network (CNN) used for image classification (top) and a Fully Convolutional Network (FCN) that is used for dense prediction (bottom). Note that FCNs do not contain any fully connected layers.}
\label{fig:FCN}
\end{figure}

Significant progress in semantic segmentation was achieved with the introduction of FCNs~\cite{long2015fully} that use the entire image to infer the dense prediction. FCN is composed of only stacked convolutional and pooling layers, without any fully connected layers, as shown in Figure \ref{fig:FCN}. Originally, the proposed network used a stack of convolutional layers without any downsampling to learn the mapping from pixels to pixels directly by maintaining the input resolution. To compensate for this and to learn sufficient representative features, they assemble multiple consecutive convolutional layers. This initial approach was able to generate impressive results but at a very high computational cost. Therefore, the model was inefficient and it's scalability was limited.

As a solution to this problem, they presented an encoder-decoder architecture. The encoder is a typical CNN pre-trained for classification tasks such as AlexNet~\cite{krizhevsky2017imagenet} or VGGNet~\cite{simonyan2014very}. The goal of the downsampling layers in the encoder is to capture deep contextual information that corresponds to the semantics. For its part, the decoder network is composed of deconvolutional and up-sampling layers. Its goal is to convert a low-resolution feature representation to a high-resolution image, recovering the spatial information and enabling precise localization, thereby yielding the dense classification.

Towards this direction, a deeper deconvolution network consisting of stacked deconvolution and unpooling layers is proposed in \cite{noh2015learning}. Oliveira~\textit{et~al.}~\cite{valada2016deep} employed a similar strategy called UpNet for part segmentation and tackled two problems: occluded parts and over-fitting. Similarly, an encoder-decoder architecture was proposed in \cite{badrinarayanan2017segnet} where the feature maps obtained with the encoder are used as input of the decoder network that upsamples the feature maps by keeping the maxpooling indices from the corresponding encoder layer. Similarly, Liu~\textit{et al.}~\cite{liu2015parsenet} proposed an approach called ParseNet, which models global context directly. Such methods have demonstrated state-of-the-art results at the time of their introduction. Among these early deep learning models, UpNet and ParseNet achieve superior performance.

Specifically, to obtain deep features that enhance the performance, the convolutional layers learn feature maps with progressively coarser spatial resolutions, and the corresponding neurons have gradually larger receptive fields in the input image. This means that the earlier layers encode features of appearance and location, and the feature extracted with the later layers encode context and high-level semantic information. As a consequence, the two main challenges of deep convolutional neural networks in semantic segmentation were identified. First, the consecutive pooling operations or convolution striding leads to smaller feature resolution. Second, the multi-scale nature of the objects in the scene were difficult to be captured. Since the features at different resolutions encode context information at different scales, this information was exploited to enhance the representation of the multi-scale objects. In the following subsection, we present different techniques that have been proposed to address these challenges.

\subsection{Encoder Variants}

Encoders are also referred to as the backbone network in semantic segmentation architectures. The encoders used for semantic segmentation are typically based on CNNs that have been proposed for image classification. The initial semantic segmentation approaches adopted  the VGG-16~\citep{simonyan2014very}, AlexNet~\cite{krizhevsky2017imagenet}, or GoogLeNet~\citep{szegedy2015going} architectures for the encoder. Each of these encoders have achieved outstanding results in the
ImageNet ILSVRC14 and ILSVRC12 \citep{deng2009imagenet} challenges. VGG was extensively used in several semantic segmentation architectures~\citep{chen2017deeplab, liu2015semantic}. A breakthrough in semantic segmentation models was achieved with the introduction of ResNet~\citep{he2016deep}. Several semantic segmentation models that employed ResNets and its variants such as Wide ResNet~\cite{zagoruyko2016wide} and ResNeXt~\citep{xie2017aggregated} achieved state-of-the-art performance on various benchmarks. Another popular encoder architecture were the new generation of GoogLeNet models such as Inception-v2, Inception-v3~\citep{szegedy2016rethinking}, Inception-v4 and Inception-ResNet~\citep{szegedy2017inception}. More recently, semantic segmentation models that employ the EfficientNet~\cite{tan2019efficientnet} family of architectures have achieved impressive results while being computationally more efficient than the previous encoders.

\subsection{Upsampling Methods}

While employing multiple sequential convolution and pooling operations in the network leads to deep features and enhances the performance of perception networks, substantial information loss can occur in the downsampled representation of the input towards the end of the network.
This loss in information can affect the localization of features as well as details of the scene elements, such as texture or boundary information. Diverse works in this direction have been proposed to prevent or recuperate the loss in information. As a solution to this problem, \citep{noh2015learning} introduced deconvolution networks composed of sets of deconvolution and un-pooling layers. The authors apply their proposed network on individual object proposals and  combine the predicted instance-wise segmentations to generate a final semantic segmentation.

The goal of employing the upsampling operations during the decoding step is to generate the semantic segmentation output at the same resolution of the input image. Given the computational efficiency of bilinear interpolation, it was extensively used in several semantic segmentation networks~\citep{long2015fully,chen2017deeplab}. Another common method to upsample the features maps is using deconvolution or transpose convolution layers. Transpose convolution computes the reverse of the convolution operation and it can used to obtain the dense prediction in the decoder of semantic segmentation architectures~\citep{mohan2014deep, de2003image, saito2016real}.

\subsection{Techniques for Exploiting Context}

Different semantic segmentation methodologies have been proposed with the aim of exploiting semantic information, local appearance, and global context of the scene that can be extracted in early and late deep features. Several methodologies propose different strategies and architectures that fuse features in the process.

\subsubsection{Encoder-Decoder Architecture}
\begin{figure}
\centering \centerline{
\includegraphics[width=0.99\linewidth]{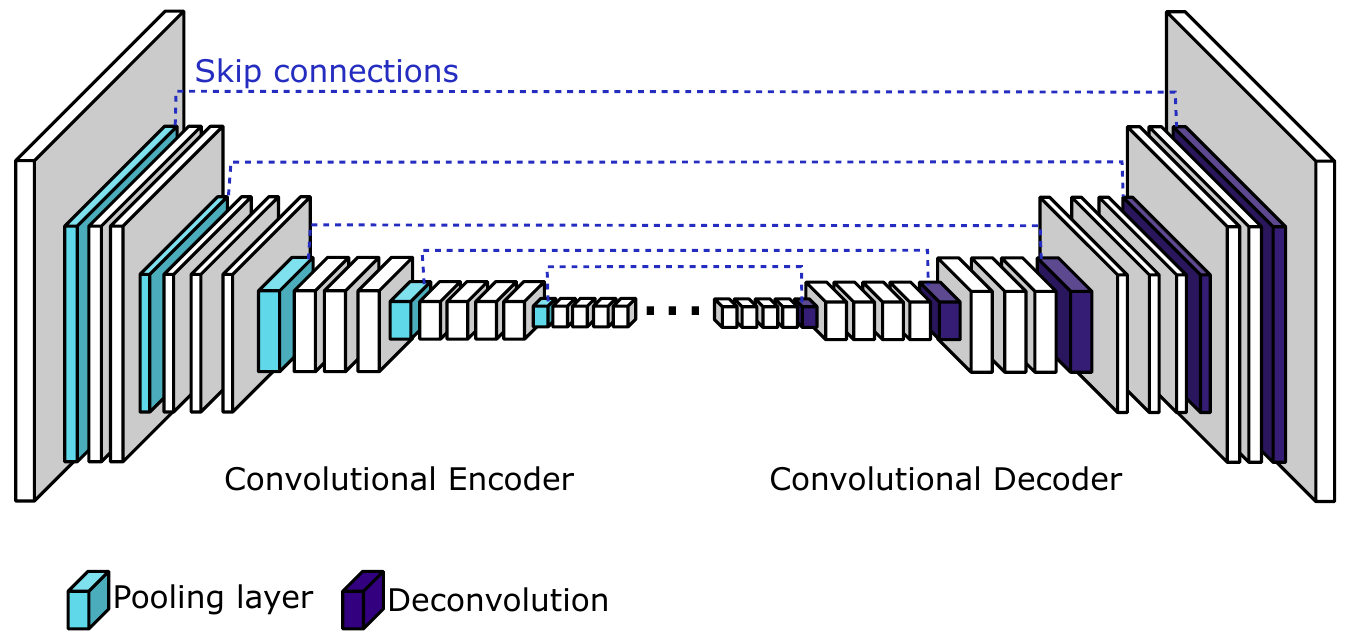}}
\caption{\it An example topology of an encoder-decoder architecture. Typically, the encoder is a pre-trained classification CNN that uses downsampling layers to capture the contextual information. On the other hand, the decoder network is composed of up-sampling layers to recover the spatial information, yielding the pixel-level classification output with the same resolution as the input image.}\label{fig:encdec}
\end{figure}

Initial semantic segmentation architectures~\cite{long2015fully, noh2015learning} used deconvolution to learn the upsampling of the encoder's outputs at low resolution. In addition, SegNet~\cite{badrinarayanan2017segnet} takes the encoder's pooling indices and later use them in the decoder to learn additional convolutional layers with the aim of densifying the feature responses. The U-Net architecture~\cite{ronneberger2015u} implements skip connections between the encoder and the corresponding decoder layers as shown in Figure~\ref{fig:encdec}. The skip connections connect encoder layers to the decoder, and allow directly transferring the information to deeper layers of the network. This network employs symmetry by increasing the size of the decoder to match the encoder. More recent encoder-decoder approaches~\cite{peng2017large,pohlen2017full, amirul2017gated,romera2017erfnet} have demonstrated the effectiveness of this structure on several semantic segmentation benchmarks.


\subsubsection{Image Pyramid}

\begin{figure}
\centering \centerline{
\includegraphics[width=0.99\linewidth]{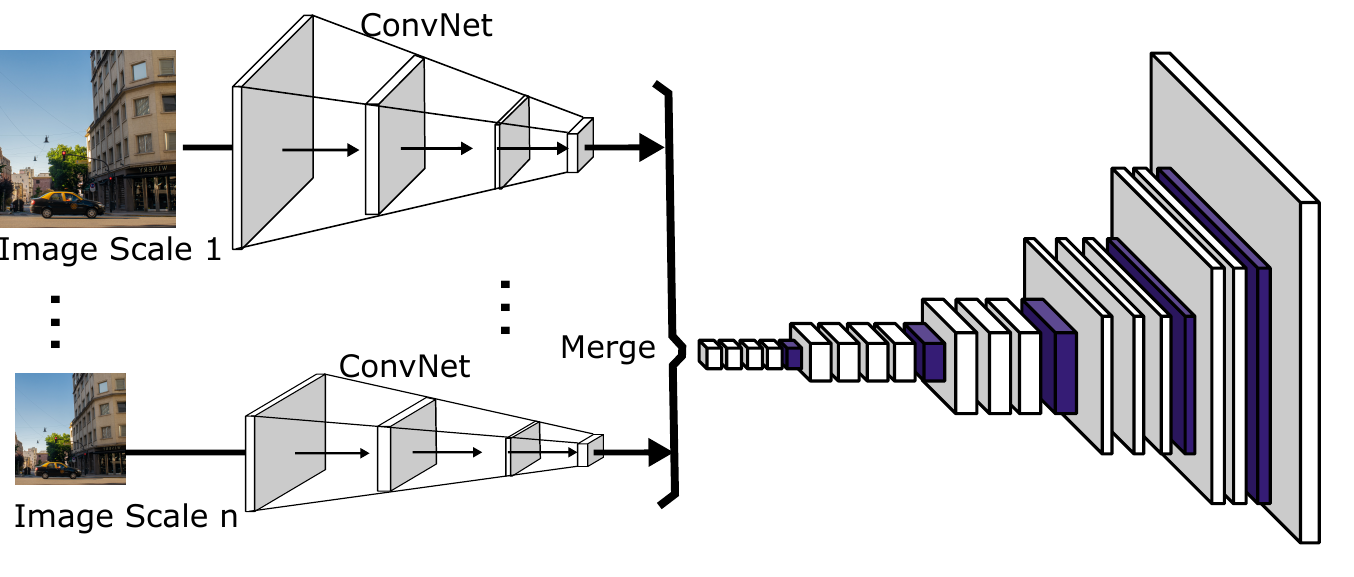}}
\caption{\it Topology of the image pyramid architecture. This network uses the same model to process the same input at a different scales. The different scales allow the network to encode different context in the image.}\label{fig:imgPyramid}
\end{figure}

The main idea with this network topology is that the same model is used to process multi-scale inputs. We present the general image pyramid network in Figure \ref{fig:imgPyramid}. Typically, the model weights are shared for multiple inputs, while the different size inputs have different purposes. Features corresponding to small scale inputs encode the wider context, and the details from small elements are encoded and preserved by the large scale inputs. Examples of this image pyramid structure include the Laplacian pyramid used to transform the input. With this transformation, each input scale is subsequently used to feed a CNN, and finally, all feature maps scales are fused together~\cite{farabet2012learning}. Other methodologies directly resize the input to different scales and later fuse the obtained features~\cite{chen2017deeplab, chen2016attention, lin2016efficient}. However, the main restriction of this models is related to limited GPU memory, given that networks that requires deeper CNNs such as \cite{he2016deep,wu2019wider,zagoruyko2016wide} which are computationally expensive.

\subsubsection{Conditional Random Fields}

Graphical models, especially Conditional Random Fields (CRFs) have been used as refinement layers in deep semantic segmentation architectures. The main objective is to capture low-level detail in regions where class intersections occur. These boundary regions are particularly difficult to segment with precision. This strategy includes additional modules placed consecutively to represent the longer-range context. To do so, a popular methodology is to integrate CRF into CNNs. Diverse methodologies have been presented as a refinement layer, including Convolutional CRFs~\cite{teichmann2018convolutional} and Dense CRF~\cite{krahenbuhl2011efficient}. Other methodologies have been proposed to train both the CRF and CNN jointly~\cite{chen2014semantic,chen2017deeplab}. These methods that incorporate CRFs have demonstrated their benefits to capture contextual knowledge and exploit finer details to enhance the class label localization in the pixels.

\subsubsection{Spatial Pyramid Pooling}

\begin{figure}
\centering \centerline{
\includegraphics[width=0.99\linewidth]{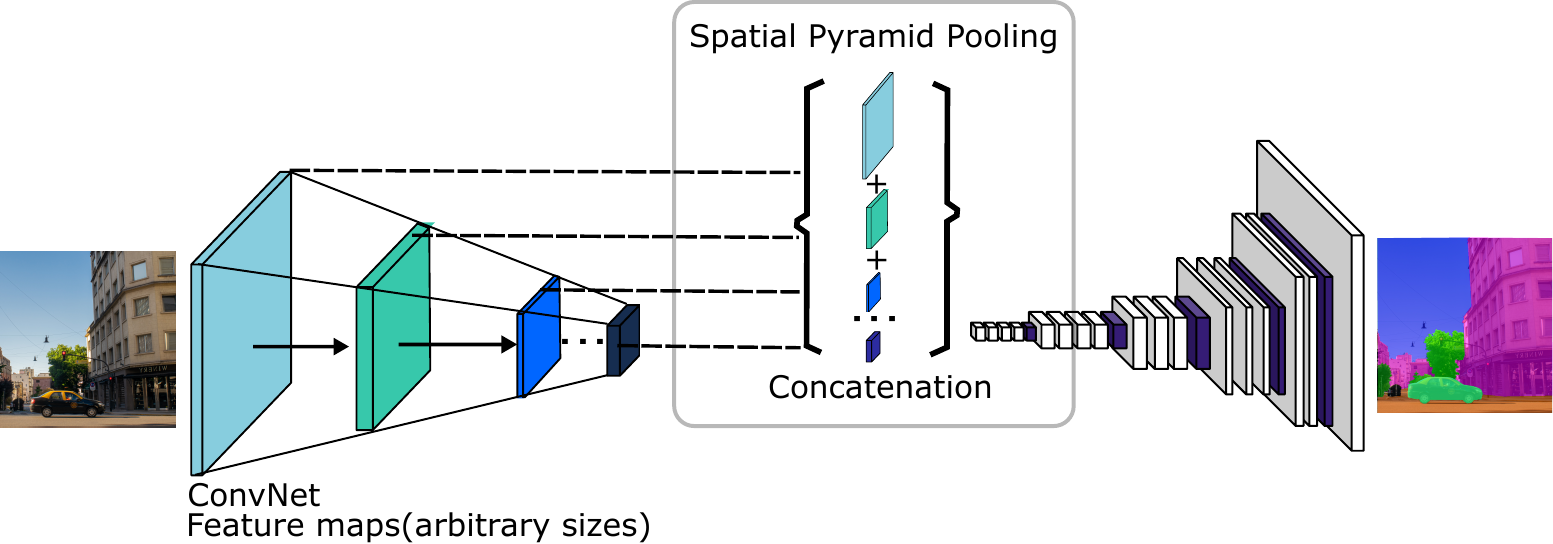}}
\caption{\it Topology of Spatial Pyramid Pooling architecture used to exploit the context found at multiple scales. This network includes a new module for multi-level pooling between the convolutional and fully-connected layers. The multi-level pooling allows this network to be more robust to the variations in object scale and deformation.}
\label{fig:SSP}
\end{figure}

This structure uses spatial pyramid pooling to obtain context at multiple scales, and it is graphically described in Figure \ref{fig:SSP}. Towards this direction, ParseNet~\cite{liu2015parsenet} exploits image-level features. Another methodology is presented in DeepLabv2~\cite{chen2017deeplab}, which uses atrous spatial pyramid pooling (ASPP) that includes parallel atrous convolution layers with different rates to consider objects at different scales. This strategy improves the segmentation performance. Following this direction, a subsequent work also proposes to generate multi-scale features that cover a larger scale range densely using the Densely connected Atrous Spatial Pyramid Pooling (DenseASPP)~\cite{yang2018denseaspp}. Furthermore, an efficient variant of the ASPP is proposed in Adapnet++~\cite{valada2019self} which captures a larger effective receptive field while decreasing the required parameters by 87\% using cascaded and parallel atrous convolutions. Additionally, in Pyramid Scene Parsing Network (PSPNet)~\cite{zhao2017pyramid}, the authors exploit the global context information by aggregating multiple region-based contexts with a proposed pyramid pooling module. PSPNet demonstrated significant improvement on several semantic segmentation benchmarks. 

\subsubsection{Dilated Convolution}

\begin{figure}
\centering \centerline{
\includegraphics[width=0.6\linewidth]{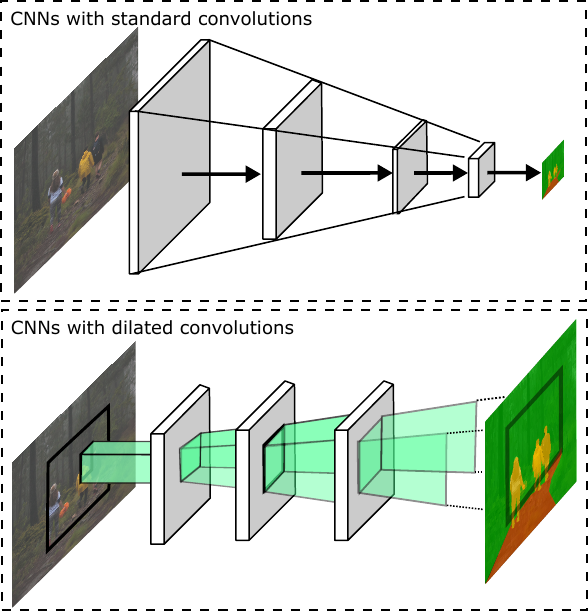}}
\caption{\it Illustration of CNNs with standard convolutions (top) and CNNs with dilated convolutions (bottom). The receptive fields of consecutive deep convolution layers output smaller resolution feature maps. The receptive field of dilated convolution grows more rapidly than in contiguous convolutional filters.}\label{fig:dilatedconv}
\end{figure}

Dilated convolutions, graphically presented in \ref{fig:dilatedconv}, are also known as atrous convolutions~\cite{chen2017deeplab} and aim to have an effective receptive field that grows more rapidly than in contiguous convolutional filters. These convolutions are an effective strategy to preserve the feature map resolution and extract deep features without using pooling or subsampling. Nevertheless, since the feature map resolutions are not reduced with the progression of the network hierarchy, using dilated convolutions requires higher GPU storage and computation. Some of the explored models that use atrous convolutions for semantic segmentation include \cite{dai2017deformable,wang2018understanding,wu2016bridging}.

\subsection{Real-Time Architectures}

Many of the architectures and topologies presented so far in this chapter typically require high computational capacity and are not efficient for real-time applications. As these architectures employ large networks for their encoder such as GoogleNet~\citep{szegedy2015going} and ResNet~\citep{he2016deep}, or use large CNN structures in different stages of the architecture, they achieve high performance but with low efficiency in terms of computation cost and runtime. To address this problem, different works propose approaches that are more suitable for real-time applications. For example, Efficient Neural Network (Enet)~\cite{paszke2016enet} is a lightweight architecture in  which the last stage of the model is removed to optimize the network to obtain a faster inference time. The main drawback of this architecture is that excluding the downsampling operations in the last stage of the network makes it unable to cover large objects since the receptive field is smaller.

As an alternative, in ERFNet~\cite{romera2017erfnet} a layer is designed to use residual connections and depthwise separable convolution with the aim of increasing the receptive field. On the other hand, in the spatial pyramid pooling structure, the convolutional feature maps are re-sampled at the same scale before the classification layer, resulting in a computationally expensive process. To tackle this problem, ESPNet is proposed as an efficient network structure~\citep{mehta2018espnet}. This approach aims to efficiently exploit both context and spatial features by incorporating an efficient convolutional module called ESP. ESPNet architecture is able to preserve the segmentation accuracy while being fast and small while requiring low power and low latency. ESPNet decomposes a standard convolution into two steps. First, point-wise convolutions to lessen the computation effort. Second, a spatial pyramid of dilated convolutions that re-samples the feature maps so that the network can learn the representations from a large effective receptive field.

The approaches mentioned above provide lightweight architectures but compromise the model accuracy. Other models such as BiSeNet~\cite{yu2018bisenet}, and LiteSeg~\cite{emara2019liteseg} explore strategies to improve computational efficiency while maintaining high accuracy. In BiSeNet, the authors design two different streams. First, the Spatial Path generates high-resolution features by using a small stride. Second, the Context Path is designed to obtain an adequate receptive field with a fast downsampling approach. Later, an additional Feature Fusion Module is employed to combine both features. In LiteSeg, the authors propose a deeper version of ASPP and use dilated convolutions and depthwise separable convolutions. As a result, LiteSeg is a faster and efficient model which provides high accuracy.

\subsection{Object Detection-based Methods}

In object detection in scenes with multiple elements, the main goal is to generate a bounding box indicating each object. Nevertheless, since the input image can include a diverse number of objects or may not have any objects, the number of objects that should be detected cannot be fixed. Therefore,  this task can not be solved using a standard CNNs followed by a fully connected layer with a predefined number of output classes. A straightforward approach to tackle this problem is to select different regions of interest from the image and employ a CNN on each region to classify the presence of a single object in the region. The architecture used to determine the regions out of the input image is called Region Proposal Network (RPN). RPNs are essential structures in the construction of algorithms that select a reduced group of regions of interest where objects might be located in the image. These algorithms aim to obtain an optimal number of region that allow the detection of all the elements in the scene, therefore reducing the required computation capacity. Some popular algorithms are YOLO~\cite{redmon2016you}, Single Shot Detector~\cite{liu2016ssd}, and Fast-RCNN~\cite{girshick2014rich} and its improved version Faster-RCNN~\cite{ren2016faster}. These networks facilitate the segmentation of the object inside a smaller region of the image. In this direction, the segmentation of instances is proposed in Mask-RCNN~\cite{he2017mask}, and YOLACT~\cite{bolya2019yolact}. They obtain a semantic segmentation output by drawing the segmentation masks and completing all the pixels of the input image.

\section{Loss Functions for Semantic Segmentation}

In this section, we discuss the most commonly employed loss functions for learning the semantic segmentation task.

\subsection{Pixel-Wise Cross Entropy Loss}

The pixel-wise cross entropy loss function~\cite{yi2004automated} inspects each pixel individually by comparing the predicted class label to the ground truth, and finally computes averages over all pixels. This loss function leads to equal learning for each pixel in the image, which can lead to problems if the image is composed of unbalanced pixel classes. With the aim of tackling class imbalance, weighting this loss for each output channel was proposed~\cite{long2015fully}. Additionally, a pixel-wise weighting loss has also been proposed~\cite{ronneberger2015u}. In this case, a larger weight is assigned to the pixels at the borders of segmented objects. The pixel-wise cross entropy loss ($\mathcal{L_{CE}}$) is computed as
\begin{equation}
\mathcal{L_{CE}}(p,y) = - \sum_c y_{o,c} \text{log} p_{o,c},
\end{equation}
where $c$ is the class label, $o$ denotes an observation, $y_{o,c} \in [0,1]$ denotes if $c$ is the correct classification for $o$, and $p$ is the predicted probability of $o$ belonging to $c$.

\subsection{Dice Loss}

The Dice loss function~\cite{milletari2016v} is based on the dice coefficient metric that is used to measure the overlap between two samples. This metric is also used to compute the similarity between a pair of images. The dice loss is computed for each class individually and then the average among class is computed to obtain a final loss. 
The dice coefficient is computed as
\begin{equation}
\text{Dice}(y, c) = \frac{2 |X \cap Y|}{|X| + |Y|}
\end{equation}
where $c$ is the class label, $X$ are the scores of each class and $Y$ is a tensor with the class labels. Based on this metric, the Dice loss ($\mathcal{L_D}$) is computed as
\begin{equation}
\mathcal{L_D}(y, c) = 1 - \text{Dice}(y, c)
\end{equation}

\section{Semantic Segmentation Using Multiple Inputs}

Thus far in this chapter, we primarily discussed techniques for semantic segmentation that take a single image as input and yield a corresponding segmentation output for that image. In this section, we further discuss semantic segmentation methods that take multiple inputs simultaneously, namely for video semantic segmentation, point cloud semantic segmentation and multimodal semantic segmentation.

\subsection{Video Semantic Segmentation}


As robots move and interact with the environment, the perception is typically dynamic with changing context and scene relationships. Most often, a temporal sequence of data is available as an input to the semantic segmentation model and we can exploit this information to enforce temporal coherence in the output. For example, if an object of a particular semantic class is present in two consecutive frames, that object should be identified with the same label across the different frames, assuring the temporal coherence of the classified pixel. To do so, different works have explored various techniques to exploit temporal information to improve the overall semantic segmentation performance~\cite{shelhamer2016clockwork, fayyaz2016stfcn, siam2017convolutional, nilsson2018semantic}. Specifically, clockwork convnets are proposed in \cite{shelhamer2016clockwork} uses clock signals to control the learning rate of different layers and a LSTM-based spatio-temporal FCN is introduced in \cite{fayyaz2016stfcn}. Similarly, \cite{siam2017convolutional} uses a spatio-temporal representation where the convolutional gated recurrent network enables learning both spatial and temporal information jointly. Furthermore, the approach presented in \cite{nilsson2018semantic} combines convolutional gated architectures and spatial transformers for video semantic segmentation.

\subsection{Point Cloud Semantic Segmentation}

\begin{figure}
\footnotesize
\centering
\setlength{\tabcolsep}{0.05cm}
{\renewcommand{\arraystretch}{0.8}
\begin{tabular}{ccc}
     \rotatebox{90}{LiDAR Semantic} & \rotatebox{90}{Segmentation} &
     \includegraphics[width=0.9\linewidth]{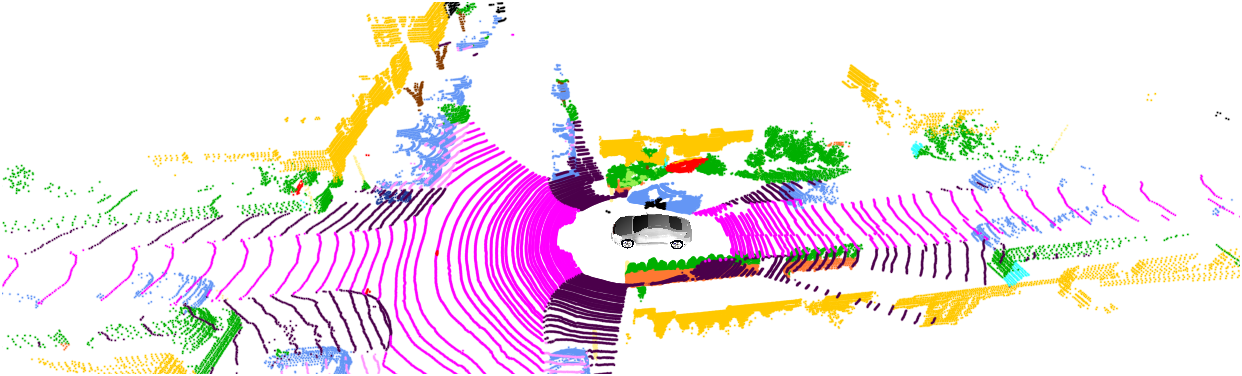}\\
     \\
     \rotatebox{90}{Spherical} & \rotatebox{90}{Projection} & \includegraphics[width=0.9\linewidth]{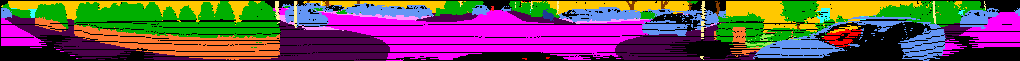} \\
\end{tabular}}
\caption{Semantic segmentation of point clouds. The top figure presents a class label assigned to each point in the 3D space. The bottom figure presents the point cloud projected into the 2D representation using spherical projections. The semantic segmentation prediction is obtained for each pixel in the projection.}
\label{fig:pc}
\end{figure}

A point cloud is a collection of points in the 3D space representing the structure of the scene. With the growing use of depth and LiDAR sensors that enable building 3D maps of the scene, methods for 3D semantic segmentation of point clouds are increasingly becoming more popular. We show in Figure \ref{fig:pc} an example of LiDAR-based semantic segmentation. We briefly discuss three main category of techniques for 3D semantic segmentation, namely point-based methods, voxel-based methods, and projection-based methods.

Point-based methods aim to process the raw point clouds directly. PointNet~\cite{qi2017pointnet} was the first approach to tackle semantic segmentation directly on the raw point clouds. The PointNet architecture allows working on unordered data with a point-wise learning methodology that uses shared multi-layer perceptrons and subsequently symmetrical pooling functions. PoinNet++~\cite{qi2017pointnet++} further extends PointNet and proposes to group points in a hierarchical manner. A similar approach~\cite{hua2018pointwise} computes per-point convolutions by grouping together neighboring points into kernel cells. In contrast, \cite{landrieu2018large} proposes to use a directed graph over the point cloud generating a set of interconnected superpoints to capture the structure and context information.


Voxel-based methods first convert the point cloud into a 3D voxel representation and them employ 3D CNN architectures on them. Voxels are volumetric discretizations of the 3D space. SegCloud~\cite{tchapmi2017segcloud} proposes to use this representation as to the input to a 3D-FCN~\cite{liu20203d}. The authors also propose a deterministic trilinear interpolation that converts the voxel predictions back to the point cloud space and subsequently employ CRFs for refinement. Nevertheless, the computation of voxels and their 3D processing consumes substantial runtime and requires very high computational cost making it unfeasible for real-time applications.

Projection-based methods project the point cloud from 3D data into 2D space to reduce the computational cost. A known transformation method is spherical projection. This approach and has been especially utilized for LIDAR semantic segmentation. The resulting 2D projection allows using 2D image methodologies for semantic segmentation, which are faster than 3D approaches. SqueezeSeg~\cite{wu2018squeezeseg} and its extension SqueezeSegV2~\cite{wu2019squeezesegv2} yield very efficient semantic segmentation result by utilizing spherical projections. Additionally, the authors in \cite{milioto2019rangenet++} uses different 2D architectures and additional post-processing stages to refine the 3D segmentation results. A more recent approach called EfficientLPS~\cite{sirohi2021efficientlps} presents a model that incorporates geometric transformations while learning features in the encoder of semantic and instance segmentation networks.

\subsection{Multimodal Semantic Segmentation}

\begin{figure}
\footnotesize
\centering
\setlength{\tabcolsep}{0.05cm}
{\renewcommand{\arraystretch}{0.8}
\begin{tabular}{cccc}
     RGB Input & Thermal Input & Depth Input & Semantic Segmentation \\
     \includegraphics[width=0.24\linewidth]{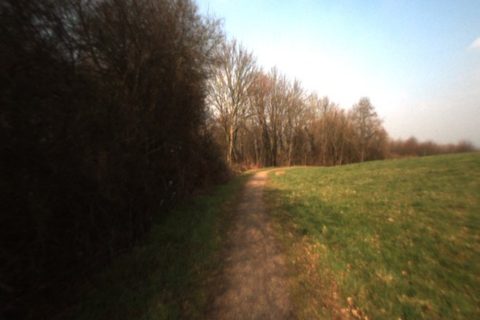} &  \includegraphics[width=0.24\linewidth]{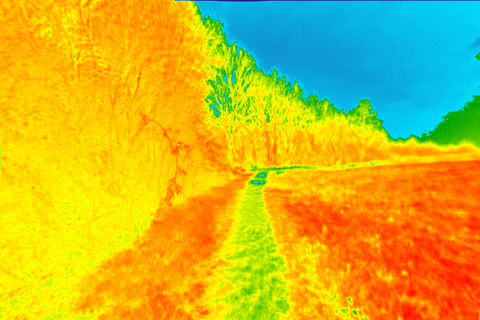}&
     \includegraphics[width=0.24\linewidth]{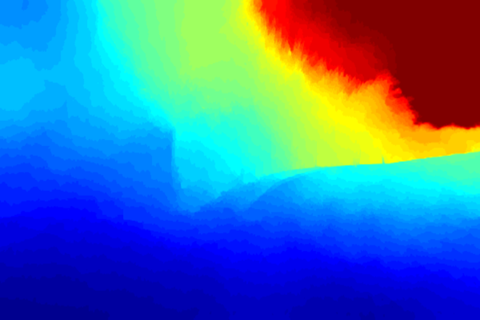}&
     \includegraphics[width=0.24\linewidth]{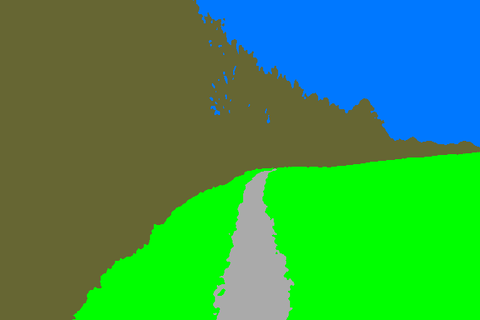}
\end{tabular}}
\caption{Example of multimodal semantic segmentation using multiple modalities as input in the Freiburg Forest dataset \cite{valada16iser}. In this case, the input modalities are RGB, Thermal, and Depth images. The semantic segmentation output is obtained by fusing the features obtained from both modalities.}
\label{fig:mm}
\end{figure}

Semantic segmentation of RGB images has led to important advances in scene understanding. However, image-based approaches suffer from visual limitations as they are susceptible to changing weather and seasonal conditions as well as varying illumination conditions. With the availability of low-cost sensors and with the goal of improving the robustness and the granularity of segmentation for robot perception, fusion of different modalities have been explored to exploit complementary features. Alternate modalities such as thermal images and depth images have been shown to be beneficial for segmenting objects in low illumination conditions by exploiting properties of objects such as reflectance and geometry.

There are three main categories of fusion techniques for multimodal semantic segmentation: early, late, and hybrid fusion. Early fusion, also known as data-level or input-level fusion, aims to combine the modalities before feeding them as input to the CNN. The obtained multimodal representation is later used as the input to a single model to exploit complementary features and leverage the interactions between low-level features of each modality. This technique typically requires that the utilized modalities have semantic similarities such as RGB and depth. A straightforward implementation of early fusion of modalities is channel stacking by concatenating multiple modalities across the channel dimension. Subsequently, a learning model can be trained end-to-end using the stacked modalities as input. In the case of RGB-Depth semantic segmentation, the first deep learning approach for multimodal early fusion proposed to concatenate the depth image with the RGB image as an additional channel~\citep{couprie2013indoor}. Later, \citep{hazirbas2016fusenet} proposed to use an encoder-decoder architecture composed of two encoder branches to extract features from RGB and depth images, and subsequently combine the obtained depth feature maps with the RGB branch.  More recently, a fusion mechanism proposed in RFBNet~\citep{deng2019rfbnet} explores the interdependencies between the encoders to provide an efficient fusion strategy. Given that early fusion approaches mainly utilize a single or unified network, they are computationally more efficient than the other techniques. Nevertheless, early fusion of modalities has its own set of limitations since it always forces the network to learn fused representations from the beginning and does it enable the network to exploit cross-modal interdependencies, and complementary features.

Late fusion techniques use individual CNN streams for each modality, followed by a fusion stage in the end which facilitates learning of further combined multimodal representations. In one of the early methods, an RGB and depth based geocentric embedding was proposed for object detection and segmentation~\cite{gupta2014learning}. To this end, the method employs two modality-specific networks to obtain feature maps that are then fed to a classifier. Other deep learning methodologies aim to map the multimodal features to a subspace \cite{eigen2013learning, jacobs1991adaptive}. In~\citep{eigen2013learning}, an adaptive gating network was proposed which generates a class-wise probability distribution over the modality-specific network streams. \cite{valada2016towards} exploits both multimodal and multispectral images in a late-fused convolution architecture. In a similar work, \citep{valada2016deep} proposes a fusion architecture that extracts multimodal features separately using individual network streams, followed by feeding the summation of the resulting feature maps through consecutive convolutional layers. Since the modality-specify streams are trained individually, the subsequent fusion training allows obtaining a combined prediction. As a result, the late fusion approach can potentially learn better complementary features along with fusion specific features. This strategy facilitates a more robust performance in cases when the features in each model have good classification performance, then the fusion further yields improvements. However, fusing feature maps towards the end of the network may not be sufficient for learning accurate, robust, and highly detailed semantic segmentation.

Hybrid fusion methodologies aim to exploit the strengths of both early and late fusion strategies. In this direction, RDFNet~\citep{park2017rdfnet} leverages the main idea behind residual learning and applies it to deep multimodal fusion for combining RGB-D features using multimodal feature fusion blocks and multi-level feature refinement blocks. On the other hand, semantics-guided multi-level fusion~\citep{li2017semantics} learns the joint feature representation in a bottom-up setup by using the cascaded semantics-guided fusion block to fuse lower-level features across modalities as a sequential model. More recently, a self-supervised model adaptation module was proposed for deep multimodal fusion~\citep{valada2019self} which dynamically adapts the fusion of semantically mature multi-scale representations by exploiting complementary cues from each modality-specific encoder.

\section{Semantic Segmentation Datasets and Benchmarks}

One important reason for the advancements and popularity of the semantic segmentation task is the availability of large-scaled labeled datasets that are publicly available. Moreover, standard benchmarks and competitions in semantic segmentation facilitate the comparison of trained models against the state of the art in different contexts, such as autonomous driving. 

Assuming that sufficient amount of labeled training data is available, it is feasible to train a semantic segmentation model from scratch for a specific application. However, the annotation process of pixel-level labeling is an arduous and expensive task. As a consequence, the amount of labeled data that is available is often insufficient. One alternative is to use semi-supervised and weakly supervised learning methods~\cite{saffar2018semantic} for annotating new datasets. Another widely used approach is transfer learning, where the model is first trained on a similar task or dataset with enough data. Transfer learning requires adaptation from the available dataset to the target dataset. A common practice is to first pre-train the model using an extensive dataset and re-train the model on the target dataset by initializing the network with the pre-trained weights. Under the premise of those pre-trained models capturing the semantic information, and therefore enabling them to train the model with less labeled samples, a model trained on an autonomous driving dataset in one city can be adapted to a different city. In these cases, the extensive and publicly available datasets and benchmarks are very helpful not only to prove and compare the capabilities of the developed methods but also as a pre-training in specific applications while using less annotated data. In this section, we briefly review public available datasets for semantic segmentation.

\subsection{Outdoor Datasets}

\begin{figure}
\footnotesize
\centering
\setlength{\tabcolsep}{0.05cm}
{\renewcommand{\arraystretch}{0.8}
\begin{tabular}{cc}
     Input RGB & Semantic Segmentation \\
     \includegraphics[width=0.5\linewidth]{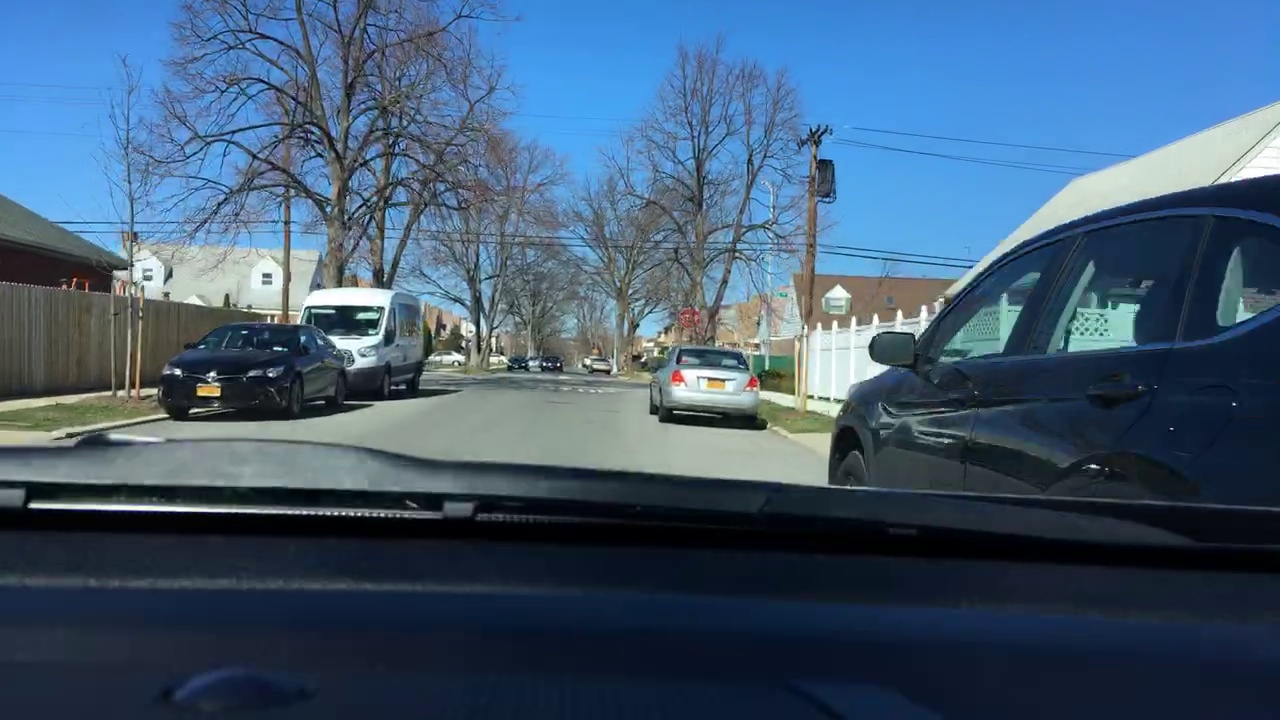} &  \includegraphics[width=0.5\linewidth]{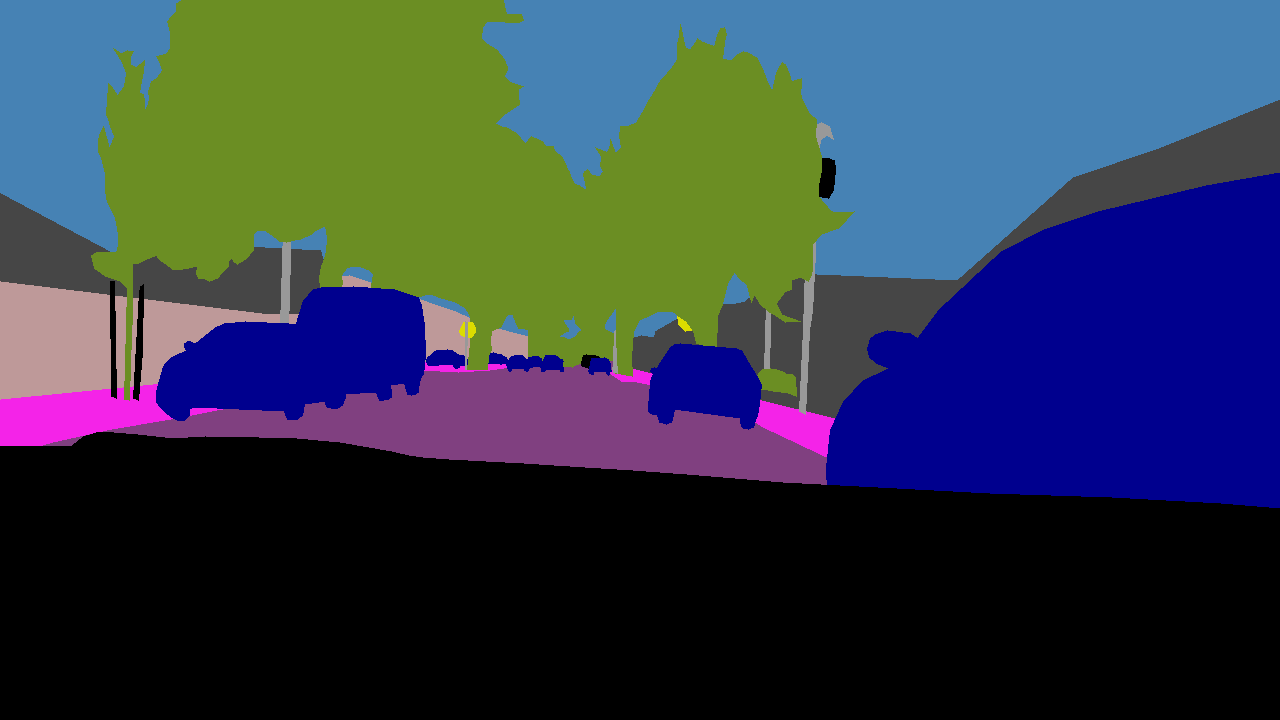} 
\end{tabular}}
\caption{An example of semantic image segmentation from the challenging BDD100K dataset. The image shows a complex urban scene with many dynamic objects.}
\label{fig:outdoor}
\end{figure}

These datasets consist of a collection of images in diverse driving scenarios including highways, densely populated urban areas, and rural areas. Many of the datasets contain challenging perceptual conditions such as high traffic density, rain, snow, fog and other seasonal pearceptual changes. In this context, the pixel labels may include pedestrians, road, car, vehicles, bicycles, and buildings. We present an example of a outdoor image in Figure \ref{fig:outdoor}.

\subsubsection{Cityscapes}

The Cityscapes~\citep{cordts2016cityscapes} dataset is a large-scale dataset consisting of urban street scenes. It is a highly diverse dataset comprising of scenes captured at different times of the day and during different seasons of the year, from over 50 European cities. The additional presence of a large number of dynamic objects further adds to its complexity making it one of the most challenging datasets for scene understand tasks. Cityscapes provides pixel-level annotations at two quality levels: fine annotations for $5,000$ images and coarse annotations for $20,000$ images. There are 30 different class labels, eight classes also have instance-specific labels. Consequently, there are three separated challenges: pixel-level semantic segmentation, instance-level semantic segmentation, and panoptic segmentation. The images in this dataset were captured at a resolution of $2048\times1024$~pixels using an automotive-grade $\SI{22}{\centi\meter}$ baseline stereo camera. The finely annotated images are divided into 2975 for training, 500 for validation and 1525 for testing. The annotations for the test set are not publicly released, rather they are only available to the online evaluation server that automatically computes the metrics and publishes the results on the leaderboard. 

\subsubsection{KITTI} 

The KITTI~\cite{Geiger2013IJRR} dataset is one of the most comprehensive datasets for autonomous driving. Besides providing groundtruth for semantic segmentation, KITTI also includes annotations for diverse tasks such as scene flow estimation, optical flow estimation, depth prediction, odometry estimation, tracking, and road lane detection. The dataset contains sequences of frames recorded with diverse sensor modalities such as high-resolution RGB, grayscale stereo cameras, and a 3D laser scanners. 

\subsubsection{Mapillary Vistas} 

The Mapillary Vistas~\cite{neuhold2017mapillary} dataset contains 25,000 high-resolution images annotated into 66 object categories. Mapillary Vistas is one of the most extensive publicly available datasets for semantic segmentation of street scenes. Additionally, this dataset presents instance-specific semantic annotations for 37 classes, and it is suitable for other scene understanding tasks such as instance segmentation and panoptic segmentation. This dataset includes diverse scenes in terms of geographic extent and conditions such as weather, season, daytime, cameras, and viewpoints. The images in this dataset range in resolutions from $1024\times768$~pixels to $4000\times6000$~pixels.

\subsubsection{BDD100K: A Large-scale Diverse Driving Video Database}

The BDD100k~\cite{seita2018bdd100k} dataset is one of the largest driving datasets consisting of diverse scenes that cover different weather conditions including sunny, overcast, rainy, as well as different times of the day. The dataset consists of 100,000 videos of 4 seconds each that were captured at a resolution of $1280\times720$ pixels. BDD100k provides annotations for semantic segmentation, instance segmentation, multi-object segmentation and tracking, image tagging, lane detection, drivable area segmentation, multi-object detection and tracking, domain adaptation and imitation learning. It provides fine-grained pixel-level annotations for 40 object classes and the dataset is split into 7000 images for training, 1000 images for validation and 2000 images for testing.

\subsubsection{Indian Driving Dataset (IDD)}

The Indian Driving Dataset (IDD)~\citep{varma2019idd} dataset is a collection of finely annotated images of autonomous driving scenarios. Instead of focusing on data from structured environments, this dataset adds novel data from unstructured environments of scenes that do not have well-delineated infrastructures such as lanes and sidewalks. It consists of 10,000 images with resolution ranging from $720\times1280$ pixels to $1920\times1080$ pixels and finely annotated semantic segmentation labels with 34 classes. The training set contains 6993 images, while the validation and testing set have 981 and 2029 respectively.

\subsection{Indoor Datasets}

\begin{figure}
\footnotesize
\centering
\setlength{\tabcolsep}{0.05cm}
{\renewcommand{\arraystretch}{0.8}
\begin{tabular}{cc}
     Input RGB & Semantic Segmentation \\
     \includegraphics[width=0.43\linewidth]{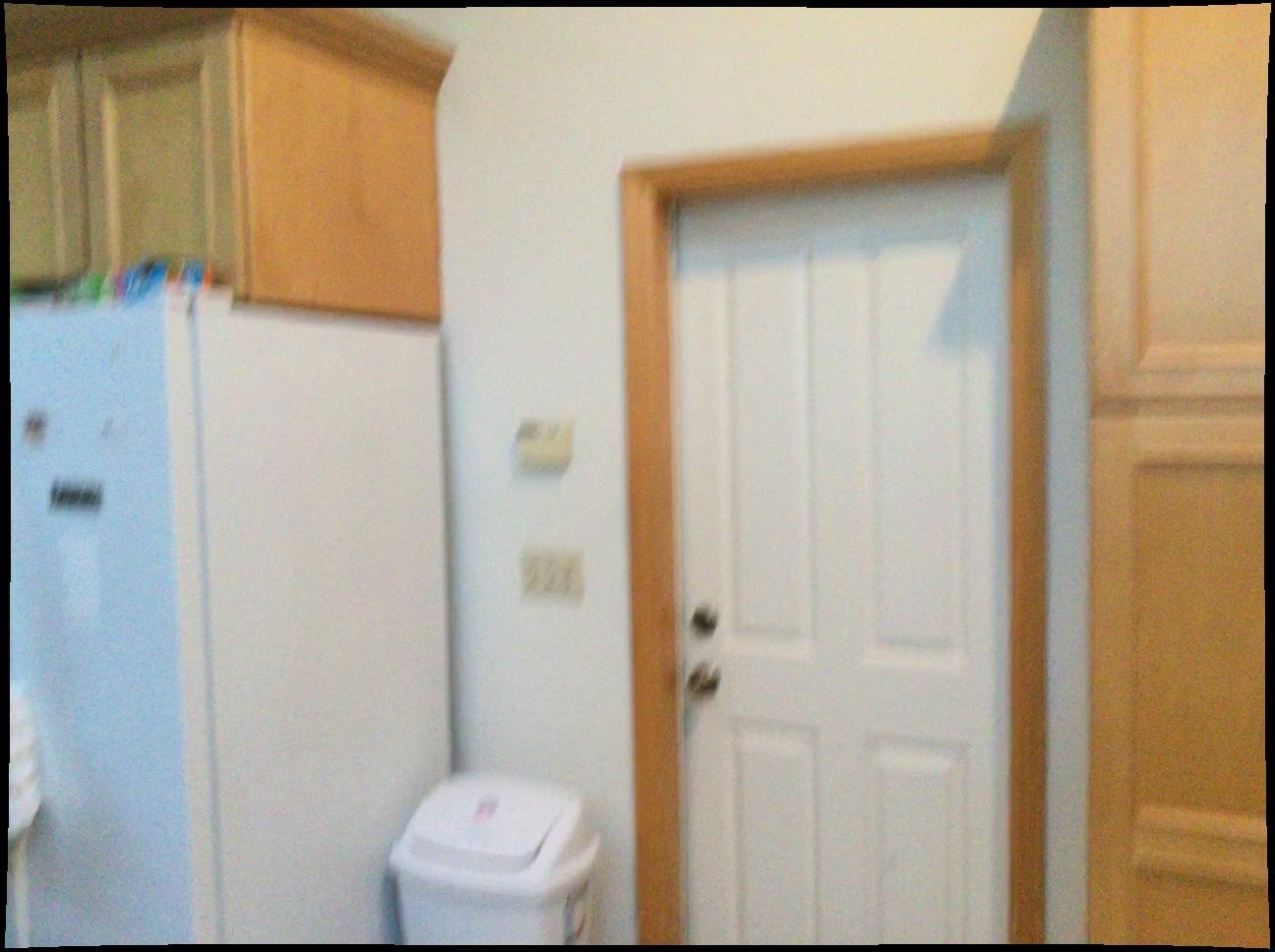} &
     \includegraphics[width=0.43\linewidth]{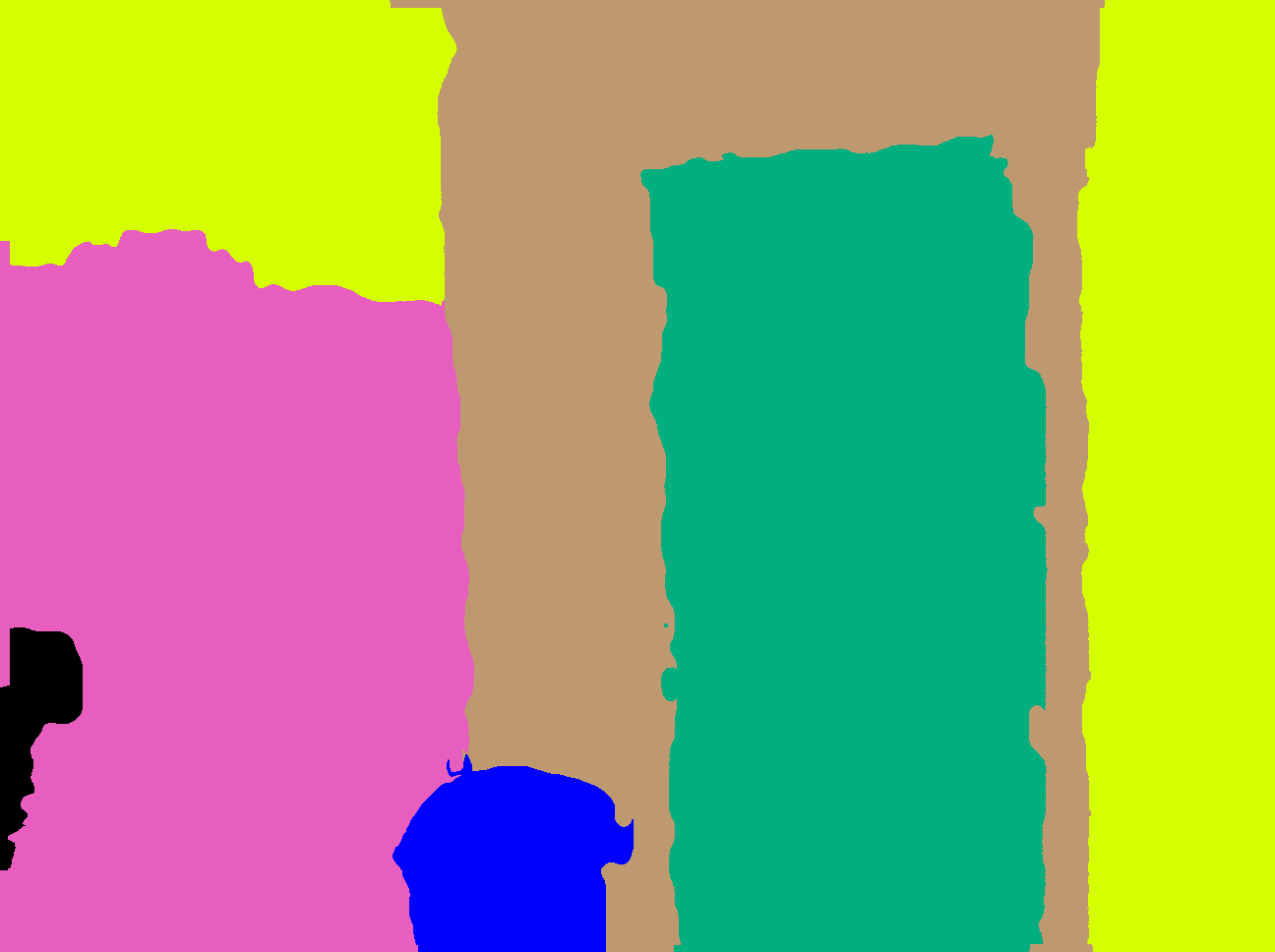} \\
 \end{tabular}}
\caption{An example semantic image segmentation from the challenging ScanNet dataset. The image shows an indoor scene where objects appear much closer than in outdoor scenes.}
\label{fig:indoor}
\end{figure}

These datasets contain indoor scenes such as offices and rooms. Besides semantic class labels for images, some of datasets also provide depth images and 3D models of the scenes. The class labels in these datasets include objects such as sofa, bookshelf, refrigerator, and bed. Additionally, indoor datasets present background class labels such as wall and floor. We present an example of a outdoor image in Figure \ref{fig:indoor}.

\subsubsection{NYU-Depth V2}

The NYU~Depth~V2 dataset \citep{Silberman:ECCV12} is a collection of video sequences recorded in indoor scenarios. The dataset was collected using both the RGB and Depth sensors from the Microsoft Kinect at 464 different scenes from three different cities. It contains $1449$ images of nearly 900 different categories with the respective dense semantic labels and additional depth information.

\subsubsection{SUN 3D}

The SUN~3D~\citep{xiao2013sun3d} dataset contains RGB-D videos of 415 sequences of 254 different indoor scenes. The sequences were captured in over 41 buildings. Only 8 sequences in this datasets has been annotated with semantic segmentation labels. In addition to the RGB-D images and the dataset also provides the camera poses for each frame.

\subsubsection{SUN RGB-D}

SUN RGBD~\citep{song2015sun} is a scene understanding benchmark suite. The dataset contains $10,335$ pairs of RGB-D images with pixel-wise semantic annotation for both 2D and 3D, for both objects and rooms. The dataset provides class labels for six important recognition tasks: semantic segmentation, object classification, object detection, context reasoning, mid-level recognition, and surface orientation and room layout estimation.

\subsubsection{ScanNet}

The ScanNet~\citep{dai2017scannet} dataset is an large scale RGB-D video dataset. It consist of 2.5 million views. Besides providing annotations for semantic segmentation, ScanNet also presents labels for 3D camera poses, surface reconstructions, and instance-level semantic segmentation.

\subsection{General Purpose Datasets}
 \begin{figure}
\footnotesize
\centering
\setlength{\tabcolsep}{0.05cm}
{\renewcommand{\arraystretch}{0.8}
\begin{tabular}{cc}
     Input RGB & Semantic Segmentation \\
     \includegraphics[width=0.43\linewidth]{Figures/Datasets/coco_rgb.jpg} &
     \includegraphics[width=0.43\linewidth]{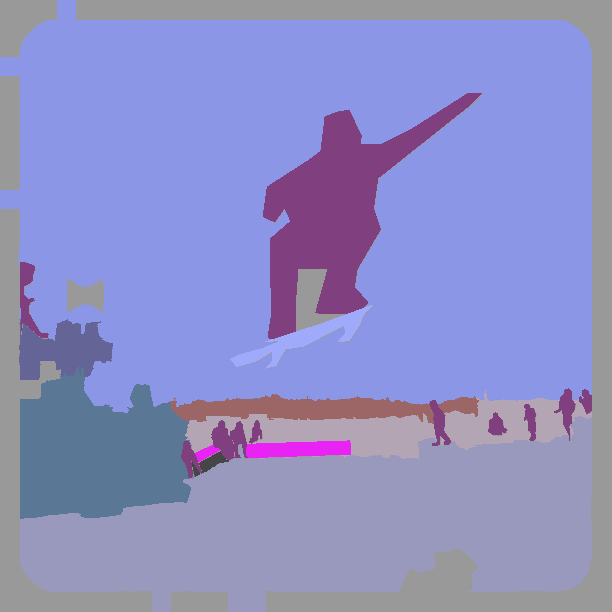} \\
\end{tabular}}
\caption{An example semantic image segmentation from the Microsoft Common Objects in Context dataset. The image shows a scene with a person playing sports. The class labels in the image include person, sky, and floor.}
\label{fig:gp}
\end{figure}
These datasets contain generic class labels including almost every type of object or background. Some of these datasets are the most standard benchmarks to measure progress in the semantic segmentation task as a whole. An example of this datasets is presented in Figure \ref{fig:gp}

\subsubsection{PASCAL Visual Object Classes (VOC)}

PASCAL Visual Object Classes \citep{everingham2010pascal} presents all types of indoor and outdoor images with 20 foreground object classes and one background class with $1,464$ images for training, $1,449$ images for validation, and $1,456$ test image. The test set is not public and is accessible only for the challenge. Besides including pixel-level annotations for semantic segmentation, this dataset also presents annotations for classification, detection, action classification, and person layout tasks.

\subsubsection{Microsoft Common Objects in Context (MS COCO)}

MS~COCO~\cite{lin2014microsoft} is an extensive large scale dataset for object detection, semantic segmentation, and captioning image set. It contains 330,000 images of complex everyday scenes with common objects. The dataset contains 200,000 labeled images with 1.5 million object instances and 80 object categories. The dataset is split into 82,000 images for training, 40500 images for validation and 80,000 for testing.

\subsubsection{ADE20K}

The ADE20K~\cite{zhou2019semantic} dataset contains more than 20,000 scenes with objects and object parts annotations composing 150 semantic categories. The average image size of the samples in the dataset is 1.3M pixels that can be up to $2400\times1800$ pixels. The dataset is split into 20,000 images for training and 2000 images for validation. Additionally, ADE20K also provides a public leaderboard.

\section{Semantic Segmentation Metrics}

Two principal criteria are usually considered during the evaluation of semantic segmentation models. The first being accuracy, which is related to the effectiveness and represents how successful the model is. The second corresponds to the computational complexity and is associated with the scalability of the model. Both criteria are essential for robots to successfully perform tasks using scene understanding models that can be deployed in resource limited systems.

\subsection{Accuracy}

Measuring the effectiveness of a semantic segmentation can be difficult, given that it requires to measure both classification and localization in the pixel space. Different metrics have been presented to measure each individual criteria or the combination of them. 

\subsubsection{ROC-AUC}

ROC stands for the Receiver-Operator Characteristic curve. ROC measures a binary classification system's ability by utilizing the trade-off between the true-positive rate against the false-positive rate at various threshold settings. AUC stands for the area under the curve of this trade-off, and its maximum value is 1. This metric is useful in binary classification problems and is suitable in problems with balanced classes. Nevertheless, given that most semantic segmentation presents an unbalance between the classes, this evaluation metric is not considered in most recent challenges.

\subsubsection{Pixel Accuracy (PA)}

Pixel accuracy is a semantic segmentation metric that denotes the percent of pixels that are accurately classified in the image. This metric calculates the ratio between the amount of adequately classified pixels and the total number of pixels in the image as
\begin{equation}
    PA = \frac{\sum_{j=1}^{k}n_{jj}}{\sum_{j=1}^{k}t_{j}},
\end{equation}
where $n_{jj}$ is the total number of pixels both classified and labelled as class $j$. In other words, $n_{jj}$ corresponds to the total number of True Positives for class $j$. $t_j$ is the total number of pixels labelled as class $j$.

Since there are multiple classes present in semantic segmentation, the mean pixel accuracy (mPA) represents the class average accuracy as
\begin{equation}
    mPA = \frac{1}{k} \sum_{j=1}^{k} \frac{n_{jj}}{t_{j}}.
\end{equation}

PA and mPA are intuitive and interpretable metrics. However, high PA does not directly imply superior segmentation performance, especially in imbalanced class datasets. In this case, when a class dominates the image, while some other classes make up only a small portion of the image, only correct classification of the dominant class will yield a high PA.

\subsubsection{Intersection over Union}

In semantic segmentation, Intersection over Union (IoU) is the overlap area between the predicted segmentation and the ground truth divided by the area of union between the predicted segmentation and the ground truth. From the following equation, it is shown that IoU is the ratio of true positives (TP) to the sum of: false alarms know as false positives (FP), misses know as false negatives (FN) and hits know as true positives (TP). IoU value ranges between 0 and 1, the lower the value, the worse the semantic segmentation performance. The IoU metric can be computed as
\begin{equation}
    IoU = \frac{TP}{TP+FP+FN},
\end{equation}
where $FP_{ij}$ corresponds to the  number of pixels which are labelled as class $i$, but classified as class $j$. Similarly, $FN_{ji}$, the total number of pixels labelled as class $j$, but classified as class $i$, corresponding to the misses of the class $j$. In the equation, the numerator extracts the overlap area between the predicted segmentation and the ground truth. The denominator represents the area of the union, by both the predicted segmentation and the ground truth bounding box.

Similarly to PA, mIoU is computed to obtain the average per-class IoU as
\begin{equation}
    mIoU = \sum_{j=1}^{k} \frac{TP_{jj}}{TP_{ij}+FP_{ji}+FN_{jj}}, i\not j.
\end{equation}


The IoU measure is more informative than PA given that it also punishes false alarms, whereas PA does not. The IoU is a very straightforward metric and it is very effective. Therefore, IoU and its variants, is widely used as accuracy evaluation metrics in the most popular semantic segmentation challenges such as Cityscapes and VOC challenge. However, the main drawbacks of these metric are that they only consider the labeling correctness without measuring how accurate are the boundaries of the segmentation, and the significance between false positives and misses is not measured.

\subsubsection{Precision-Recall Curve (PRC)-based Metrics}

Precision refers to the ratio of successes over the summation of successes and false alarms and recall relates the successes over the summation of successes and misses. Precision and recall is computed as
\begin{align}
    Precision &= \frac{TP}{TP+FP},\\
    Recall &= \frac{TP}{TP+FN}.
\end{align}
 
PRC is used to represent the trade-off between precision and recall for binary classification. PRC has the ability of discriminating the effects between the false positives and false negatives. Therefore, PCR-based metrics are widely used for semantic segmentation.


\begin{itemize}
    \item F score or dice coefficient is very similar to the IoU and also ranges from 0 to 1, where 1 means the greatest similarity between the segmentation and ground truth. It is a normalised measure of similarity, and is defined as
    \begin{equation}
        F-score = \frac{Precision \times Recall}{Precision + Recall}.
    \end{equation}
    
    \item PRC-AuC is defined as the area under the PRC. This metric describes the precision-recall trade-off under different thresholds.
    \item Average Precision (AP) consists of a single value summarising the shape and the AUC of PRC. This is one of the most used single value metric for semantic segmentation. The mean average precision (mAP) is also presented for all classes.
\end{itemize}

\subsection{Computational Complexity}

Besides the model accuracy, computational complexity of the semantic segmentation model is critical factor to assess. Consequently, the following metrics have been used to measure the time to complete the task and the computational sources demanded by the model.

\subsubsection{Runtime}

It refers to the total time that the model requires to produce the output, starting from the image given as input until the dense semantic segmentation output is generated. This metric highly depends on the hardware that is used. Therefore, when this metric is reported, it also includes a description of the system used.

\subsubsection{Memory Usage}

Given the ample applications of semantic segmentation, memory usage is an essential metric to report. This indicates how feasible is the deployment of the perception models on devices with limited computational resources. The goal of many semantic segmentation algorithms is to obtain the best possible accuracy with limited memory. A commonly employed metric is the maximum memory required for the semantic segmentation model. 

\subsubsection{Floating Point Operations Per Second}

The Floating Point Operations Per Second (FLOPs) refers to the number of floating-point calculations per second are required. It is used to measure the complexity of the CNN model. Assume that we use a sliding window to compute the convolution and we ignore the overhead due to nonlinear computation, then the FLOPs for the convolution kernel is given by
\begin{equation}
    FLOPs = 2HW(C_{in} K^2 + 1) C_{out}
\end{equation}
where $H$, $W$, and $C_{in}$ are the height, width and number of channels in the input feature map respectively. $K$ is the size of the kernel and $C{out}$ is the number of channels in the output.



\section{Conclusion}

In this chapter, we discussed the semantic segmentation task for robot perception. We highlighted the essential role of deep learning techniques in the field, and we described the most popular datasets that can be used to train semantic segmentation models for robotics in different environments. We described related the algorithms, architectures, and different strategies that have been proposed to improve the semantic segmentation output. We have presented the required concepts, techniques, and general tools that comprise the topic. Semantic segmentation methods have a great potential as an important component to enhance perception systems of robot that operate and interact in real-world scenarios. 

A limitation found in perception methodologies based on supervised learning is the large amount of labeled data that is required and the consequent labeling process. Given that supervised learning methods rely on a massive amount of annotated data and semantic segmentation requires dense labels for classification, the collection of datasets can be arduous, expensive, and sometimes unfeasible. Nowadays, other learning techniques such as weakly-supervised, self-supervised and unsupervised learning have been recently explored. Transfer learning is a technique that allows to first train a general model using a large annotated dataset and then fine-tune the model using a reduced number of samples from the main application. Self-supervised learning aims to use details or attributes inherent to the images that can be captured without extra annotation steps. These inherent attributes can be used to initially pretrain the network and reduce the required amount of data to train the target model. These techniques represent an interesting contribution towards mitigating the dependency of training the models on a large amount of annotated data.

In the specific case of robotics, it is usually required that the employed models run in real-time and with limited computational resources. In this regard, there is still room for improvement in scene understanding. In applications where the robot is expected to rapidly react according to the conditions and situations in the environment, such as in autonomous vehicles, it is necessary to develop segmentation models with a short inference time. Developing such models without compromising on accuracy is also an interesting direction for research.

Semantic segmentation can be considered as the starting point of holistic scene representation by training models that are able to represent the complete scene, including objects and background. After semantic segmentation, other tasks that further detail the scene have been proposed. Such is the case of panoptic segmentation that combines semantic and instance segmentation. Recently, panoptic segmentation was extended to temporal sequences of frames where the instances also conserve the assigned label so that the instances are time coherent. The evolution of tasks that gradually detail the information contained in the scene highlights the importance of semantic segmentation. 

\bibliographystyle{elsarticle-num}
\bibliography{chapter12-bibliography}


\Backmatter

\end{document}